\documentclass{ecai} 
\usepackage{latexsym}
\usepackage{amssymb}
\usepackage{amsmath}
\usepackage{amsthm}
\usepackage{booktabs}
\usepackage{enumitem}
\usepackage{graphicx}
\usepackage{color}

\usepackage{soul}
\usepackage{url}
\usepackage[hidelinks]{hyperref}
\usepackage[utf8]{inputenc}
\usepackage{subcaption}

\newcommand{\BibTeX}{B\kern-.05em{\sc i\kern-.025em b}\kern-.08em\TeX}

\begin{document}

%%%%%%%%%%%%%%%%%%%%%%%%%%%%%%%%%%%%%%%%%%%%%%%%%%%%%%%%%%%%%%%%%%%%%%%%

\begin{frontmatter}

%%% Use this command to specify your submission number.
%%% In doubleblind mode, it will be printed on the first page.

\paperid{9886} 

%%% Use this command to specify the title of your paper.

\title{GeoLLaVA: Efficient Fine-Tuned Vision-Language Models for Temporal Change Detection in Remote Sensing}

%%% Use this combinations of commands to specify all authors of your 
%%% paper. Use \fnms{} and \snm{} to indicate everyone's first names 
%%% and surname. This will help the publisher with indexing the 
%%% proceedings. Please use a reasonable approximation in case your 
%%% name does not neatly split into "first names" and "surname".
%%% Specifying your ORCID digital identifier is optional. 
%%% Use the \thanks{} command to indicate one or more corresponding 
%%% authors and their email address(es). If so desired, you can specify
%%% author contributions using the \footnote{} command.

\author[A]{\fnms{Hosam}~\snm{ Elgendy}\thanks{Corresponding Author. Email: hosam.elgendy@mbzuai.ac.ae}}
\author[A]{\fnms{Ahmed}~\snm{Sharshar}}
\author[A]{\fnms{Ahmed}~\snm{Aboeitta}} 
\author[A]{\fnms{Yasser}~\snm{Ashraf}}
\author[A]{\fnms{Mohsen}~\snm{ Guizani}}

\address[A]{Mohamed bin Zayed University of Artificial Intelligence (MBZUAI), Abu Dhabi, UAE}

%%% Use this environment to include an abstract of your paper.

\begin{abstract}
    Detecting temporal changes in geographical landscapes is critical for applications like environmental monitoring and urban planning. While remote sensing data is abundant, existing vision-language models (VLMs) often fail to capture temporal dynamics effectively. This paper addresses these limitations by introducing an annotated dataset of video frame pairs to track evolving geographical patterns over time. Using fine-tuning techniques like Low-Rank Adaptation (LoRA), quantized LoRA (QLoRA), and model pruning on models such as Video-LLaVA and LLaVA-NeXT-Video, we significantly enhance VLM performance in processing remote sensing temporal changes. Results show significant improvements, with the best performance achieving a BERT score of 0.864 and ROUGE-1 score of 0.576, demonstrating superior accuracy in describing land-use transformations. Code and dataset are available through Github.
\end{abstract}

\end{frontmatter}

%%%%%%%%%%%%%%%%%%%%%%%%%%%%%%%%%%%%%%%%%%%%%%%%%%%%%%%%%%%%%%%%%%%%%%%%

\begin{figure*}[t!]
    \centering
    \includegraphics[width=0.75\textwidth]{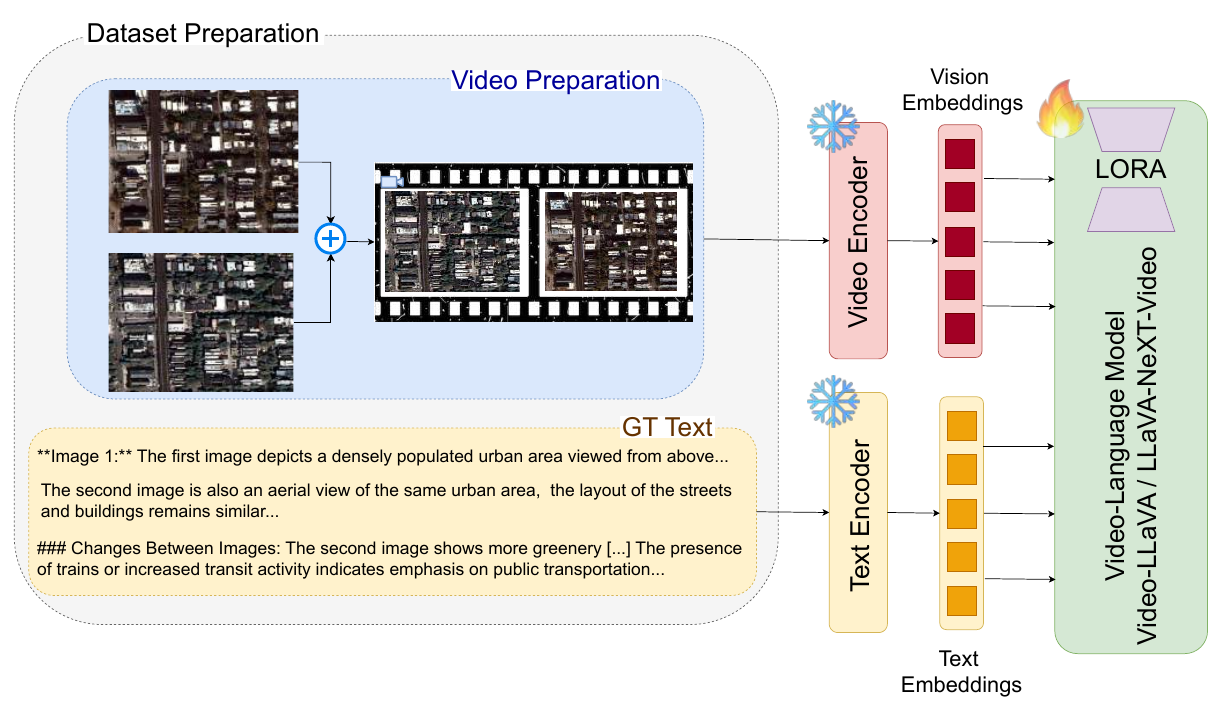}
    \caption{Overview of our pipeline using Video-LLaVA and LLaVA-NeXT-Video models. The satellite images are first aggregated and converted to a video, which is processed with the video encoder. The descriptions are processed with the text encoder, and alongside the visual embeddings, are passed to finetune the models through LoRA.}
    \label{fig:overview}
    \vspace{1em}
\end{figure*}

\section{Introduction}

Understanding temporal changes in remote sensing data is critical for numerous applications, particularly in environmental monitoring, urban planning, and geographical information systems (GIS) \cite{rev1}. Observing and interpreting how geographical features evolve over time provides insights into environmental trends, land use changes, and human impacts on the earth's surface \cite{landchange,lch1,lch2}. The advancement of Vision language models (VLMs), has made it possible to automate and enhance the detection and interpretation of such temporal changes \cite{rev2,rev3}.

Despite progress in VLMs, several limitations persist. A key challenge is their high computational demand, as training and fine-tuning large-scale models require significant resources, limiting practical use, especially for large datasets. Furthermore, many VLMs are optimized for static, unchanging scenery and struggle to capture temporal changes \cite{rev4}, which are critical in contexts such as deforestation, urban sprawl, or seasonal variation. This is further hindered by a lack of annotated remote sensing datasets focused on temporal changes, highlighting a major research gap \cite{rev5}.

This paper addresses this research gap by introducing an annotated remote sensing dataset of video frame pairs, designed to capture evolving patterns and adapt VLMs for sequential captioning. Using video frames spaced across different time intervals, the task prompts the model to describe changes between two moments \cite{rev6}, enabling it to articulate transitions or events more effectively. \footnote{The annotated data and code are available upon acceptance.}

To enhance the ability of models like Video-LLaVA and LLaVA-Next \cite{llava,liu2024llavanext}, we employ efficient fine-tuning techniques such as Low-Rank Adaptation (LoRA) \cite{lora} and Quantized LoRA (QLoRA) \cite{qlora}, applied on both few-shot and full datasets to balance accuracy and efficiency. Model pruning further reduces resource usage while maintaining performance, making these models viable for real-time applications. Our contributions are summarized as follows:
\begin{itemize}
    \item \textbf{Introduction of a Novel Dataset: }We created an annotated dataset consisting of video frame pairs that track temporal changes in geographical landscapes, particularly focused on urban and environmental transformations over time.

    \item \textbf{Optimized Fine-tuned Model: } By employing techniques like LoRA, QLoRA, and model pruning, we enhanced the efficiency and accuracy of video-language models, specifically Video-LLaVA and LLaVA-NeXT-Video, for detecting temporal changes. These models are evaluated using different metrics including, ROUGE, BLEU, and BERT.

    \item  \textbf{Comprehensive Ablation Study: } We conducted an extensive ablation study to assess the impact of different configurations, including LoRA parameters (scale ($\alpha$), rank (r)), quantization, and pruning ratios.

\end{itemize}

Although our approach does not introduce novel change detection algorithms, its primary contribution lies in the scale and diversity of the dataset encompassing a variety of classes. A distinctive aspect of our method is altering the images as videos enabling seamless integration with existing video processing pipelines. Furthermore, we provide textual descriptions and concise summaries that effectively capture visual changes, including both significant and background objects variations within satellite imagery.

\section{Related Work}

Remote sensing datasets are essential for the detailed analysis of temporal and spatial changes within dynamic environments. Foundational datasets such as LEVIR-CD \cite{cd} and FloodNet \cite{flood} have significantly advanced the field of change detection. LEVIR-CD primarily focuses on bi-temporal imagery from Google Earth to monitor urban development \cite{cd}, while FloodNet utilizes UAV-based data for assessing disaster impacts \cite{flood}. Additionally, datasets like SpaceNet and ERA have contributed to the domains of feature extraction and event recognition, respectively, whereas ISBDA offers granular disaster impact assessments \cite{SpaceNet,Era,MSNet}.

The limitations of current datasets are clear when considering LEVIR-CD's restricted scope of 637 image pairs and RSICap's focus on static scene descriptions, which fail to support studying temporal changes \cite{cd,rsgpt}. Although SkyScript boasts a substantial corpus of 2.6 million image-text pairs, it focuses on static imagery rather than evolving visual data \cite{wang}. Additionally, methods like RemoteCLIP highlight the challenges of combining visual and textual features without explicitly incorporating temporal dynamics \cite{remoteclip}.

Recent advancements in VLMs have substantially impacted remote sensing, particularly by enabling the integration of visual data with linguistic descriptions. This progress has facilitated tasks such as image captioning, zero-shot classification, and visual question answering \cite{rvlm1,rvlm2,rvlm3,rvlm4,kuckreja2023geochat}. Notable models like RemoteCLIP and GeoChat have endeavored to merge visual and textual data through training on extensive image-text datasets \cite{remoteclip,kuckreja2023geochat}. However, their applicability in remote sensing remains constrained due to a predominant focus on static image datasets, which provide limited description to the visible objects in the image and lacks temporal dependencies critical to multi-temporal data analysis.

Despite efforts to enhance these models using specialized datasets like RSICap, VLMs continue to exhibit limitations in effectively capturing and analyzing temporal changes, a capability fundamental to environmental monitoring and urban development applications \cite{wang}. GeoChat, while improving multitask conversational capabilities within remote sensing, still lacks the necessary capabilities to evaluate image evolution over time \cite{kuckreja2023geochat}. Additionally,  RemoteCLIP has successfully integrated multi-modality for various computational tasks, their functions remain predominantly limited to zero-shot classification, without extending it to temporal scene analysis \cite{remoteclip}. 

Finally, recent efforts that involved a multi-step approach to describe images for the same location over time were explored and demonstrated promising results \cite{tsujimoto2024towards,guo2024skysense}, but were limited by not processing images simultaneously, or focusing on segmentation of the changes, rather than a descriptive summary.

\section{Dataset Introduction and Processing}

To enable visual-language models (VLMs) to process temporal information, we propose a large-scale dataset comprising scene descriptions and change detection for training VLM architectures. This dataset includes visual interpretations of each image and summaries of changes between image pairs, providing insights into transformations in nature and civilization over time.

\subsection{fMoW Dataset} The fMoW RGB dataset, introduced in \cite{fmow}, is a high-resolution satellite imagery dataset targeting the classification of 62 categories \cite{satmae}. It consists of 363,571 training images and 53,041 validation images, with all multi-spectral imagery converted to JPEG format for ease of use. Images were acquired globally between 2002 and 2017, and the dataset was published in 2018. Its high spatial resolution, ranging from 0.3 to 3.7 meters, enhances the accuracy of change detection and descriptions of natural and urban environments compared to other sources.

\subsection{Creating Image Pairs} 
\begin{figure*}[ht] \centering \includegraphics[width=\textwidth]{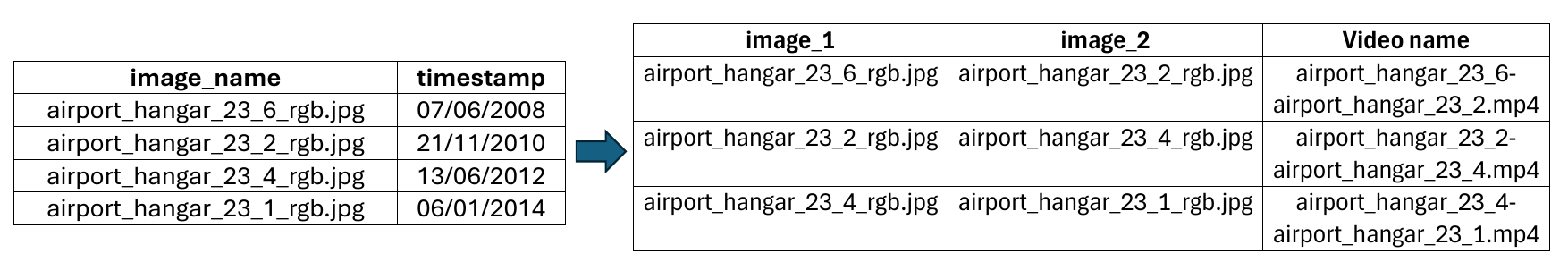} \caption{Overview of the video creation process from the original fMoW dataset images.} \label{fig:creation}
% \vspace{0.5em}
\end{figure*}

Using metadata, we sorted images by location and timestamp to create an ordered list based on the unique $"location\_id"$ identifiers. For each location, we selected image pairs that are at least 12 months apart. For example, starting with $image\_1$, we find $image\_2$ as the next image satisfying the time difference, and this process continues sequentially from $image\_2$, as illustrated in Figure \ref{fig:creation}.

Due to the size and processing limitations of ChatGPT \cite{openai2024chatgpt}, images larger than 1MB were excluded from the training and validation datasets, resulting in the removal of 5,379 training images (1.4\%) and 785 validation images (1.4\%).

\subsection{Data Splits} After filtering and annotation, we created a dataset of 100,000 image pairs from 173,348 images for training and 6,042 pairs from 11,349 images for testing. The test set was derived from the original fMoW validation set, with some images randomly selected for training to complete the 100,000 pairs. These splits are provided to ensure reproducibility of model results. While annotation costs with ChatGPT were a consideration, the dataset sufficiently meets the project’s objectives. The final dataset retains most original classes and includes image resolutions ranging from (293,230) to (4766,4634) pixels.

\begin{figure*}[ht!]
    \centering
    \begin{subfigure}{0.48\textwidth}
        \centering
        \includegraphics[width=\linewidth]{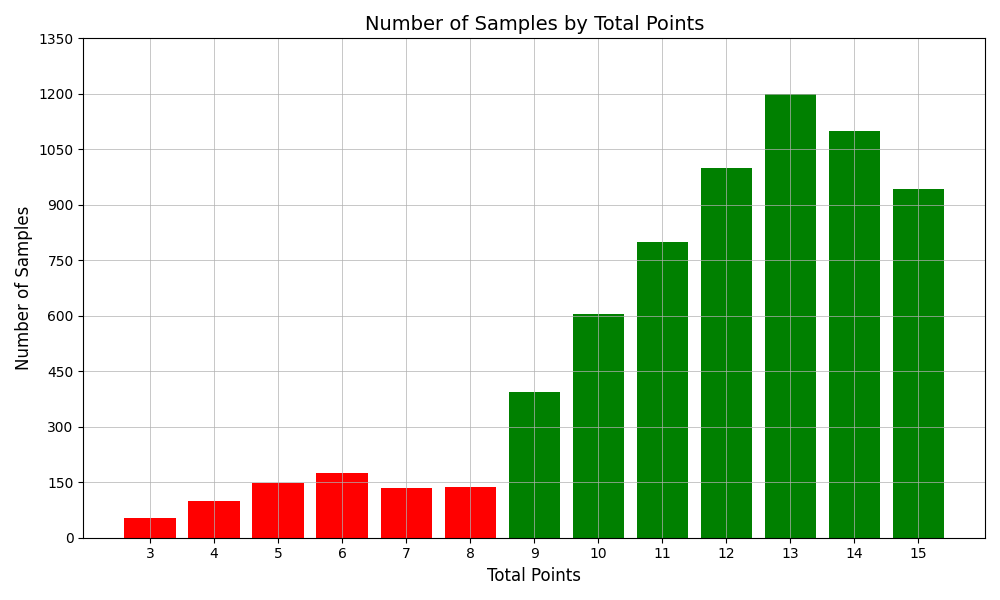}
        \caption{Total distribution of samples by score.}
        \label{fig:total_distribution}
    \end{subfigure}
    \hfill
    \begin{subfigure}{0.48\textwidth}
        \centering
        \includegraphics[width=\linewidth]{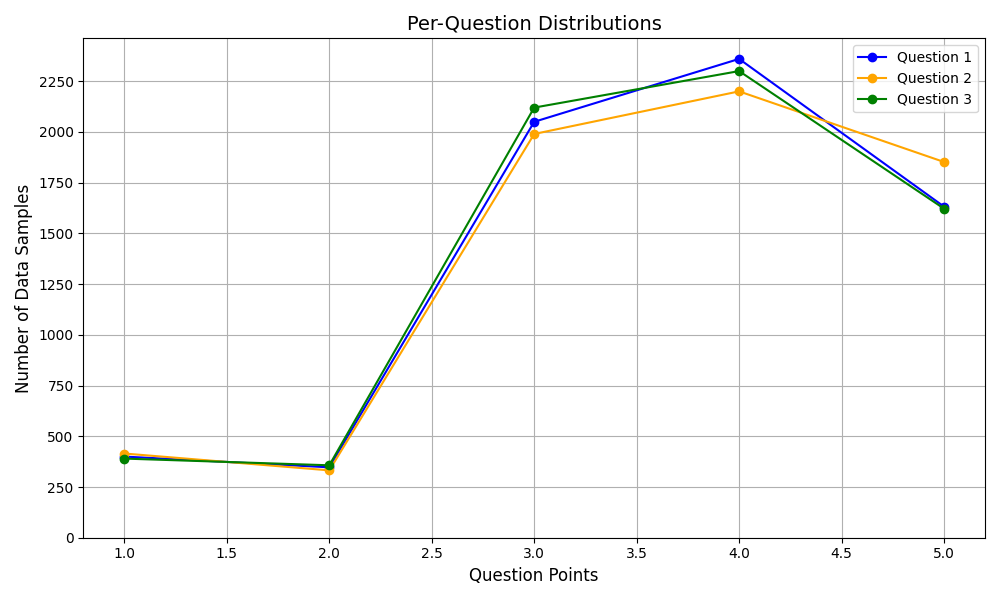}
        \caption{Distribution of points scored per question in the criteria.}
        \label{fig:per_question_distribution}
    \end{subfigure}

    \vspace{1.5em}  
    \caption{Overall sample and per-question distributions.}
    \label{fig:visualized_distribution}
    \vspace{0.5em}
\end{figure*}

It is worth noting that we manually reviewed the testing dataset to verify the authenticity and accuracy of GPT model annotations prior to model assessment. Questions were presented to the annotators (authors) when reviewing the samples, asking about the accuracy of descriptions in terms of key objects and visible details. The total set was split among the annotators, with no overlap due to the subjectivity behind rating the samples, eliminating the inter-rater overlap. By further checking the major visible object in the image, it is possible to compare it to the ground truth class in the dataset, for further assessment. Figures \ref{fig:total_distribution} and \ref{fig:per_question_distribution} showcase the distribution of the images and their rating, where samples with a score of 9 or above were kept in the dataset.

\subsection{Temporal Annotations} With image pairs established, annotations are necessary to describe each satellite image and summarize the changes between them relative to the time difference. This was achieved using OpenAI's API, where each image was processed through the "GPT-4o mini" model with the prompt:

\begin{quote} \textit{"Briefly describe each image independently, then explain the changes happening between them."} \end{quote}

The API responses typically provide descriptions of objects and landscapes, including water bodies, green ecosystems, and urban areas. The second image description often references the first, highlighting structural and environmental changes. The final paragraph explicitly details the changes, focusing on seasonal variations, vegetation dynamics, and alterations in urban landscapes.

Annotations for training and testing were generated in the standard LLaVA format \cite{llava}, including a unique ID, video name, and conversational data between "human" and AI. This data is structured from multiple templates for the human prompts, with API responses serving as "GPT" replies. The conversational format is structured as:

\begin{center}
HUMAN: \textless{}Question/Prompt\textgreater{} \textless{}Video-tokens\textgreater{} \\
GPT: \textless{}Image Descriptions and Changes Summary\textgreater{} \\
\end{center}

In conclusion, our dataset provides sufficient training samples for fine-tuning VLMs and is derived from open-source data. The annotations created contribute to making the largest available dataset for grounded image captioning of satellite images while remaining open-source and accessible.

\section{Experimental Setup}
This work presents an efficient and optimized fine-tuning pipeline aimed at enhancing the temporal understanding of geographical landscapes while highlighting significant land-use changes. We propose an architecture that integrates video processing, custom prompt construction, and fine-tuning tailored for state-of-the-art (SOTA) VLMs. Each segment of the architecture contributes uniquely to the overall system's functionality.

Initially, pairs of images are transformed into videos, with each image serving as an individual frame. These videos are then processed through a video encoder, which uniformly samples the frames and outputs a tensor array of visual information. Simultaneously, the corresponding text inputs pass to the VLM for textual encoding, enabling the model to align the textual and visual data effectively. Through fine-tuning, the model parameters are updated, transitioning from a general-purpose model to a specialized domain-specific model capable of accurately describing the input frames and detecting changes in accordance with the provided prompts. An overview of this complete system is illustrated in Figure \ref{fig:overview}.

The fine-tuning process facilitates efficient learning and optimization of the model parameters. To assess the model’s capability within the specific domain, we conducted zero-shot tests using the chosen base models, demonstrating their ability to perform well on unseen data. Additionally, a 10k sampled sub-dataset was used for few-shot tuning, allowing for targeted adjustments based on specific examples of land-use changes.

\subsection{Model Fine-tuning}
Pre-training VLMs is typically computationally intensive and time-consuming. Consequently, fine-tuning presents an effective alternative that preserves most of the model's parameters while enhancing performance on downstream tasks. Fine-tuned models can often outperform the original general models, utilizing fewer computing resources and requiring less training time \cite{patil2024review}. This advantage motivates the use of Parameter-Efficient Fine-Tuning (PEFT) methods for tasks involving geographical change detection.

In our work, we focus on fine-tuning two distinct models that have demonstrated a robust understanding of temporal data through video processing within the VLM framework for question-answering and captioning. The first model, LLaVA-NeXT \cite{liu2024llavanext}, was introduced in early 2024, offering improved reasoning and world knowledge compared to other large models. It exhibits data efficiency comparable to SOTA models such as LLaVA-1.5 \cite{llava15}, while delivering higher image resolution and enhanced visual conversation capabilities. Shortly after the release of LLaVA-NeXT, a video variant was introduced, named LLaVA-NeXT-Video, which has demonstrated strong performance in zero-shot video tasks.

The second model utilized for comparison is Video-LLaVA \cite{videollava}, which excels in understanding visual language for downstream tasks and surpasses many existing video language models across various benchmarks. 
Both projects have multiple variations based on the number of parameters for the models. For simplicity, we have chosen to use the 7B parameter variation from both models. The 7B variations can be fine-tuned with PEFT techniques on a single GPU, making them particularly well-suited for our dataset.

\subsection{Low Rank Adaptation}
Low Rank Adaptation (LoRA) is based on a pivotal insight that the disparity between the fine-tuned weights for a specific task and the original pre-trained weights often exhibits “low intrinsic rank”, which implies that the disparity can be approximated by a matrix of low rank \cite{lora}. 

For an initial pre-trained weight matrix \(W_0 \in \mathbb{R}^{d \times k}\), LoRA limits its update through a low-rank decomposition \(W_0 + \Delta W = W_0 + BA\), where \(B \in \mathbb{R}^{d \times r}\), \(A \in \mathbb{R}^{r \times k}\), and the rank \(r \ll \min(d, k)\). Throughout the training process, \(W_0\) remains unchanged and does not receive gradient updates, whereas A and B are endowed with trainable parameters. It is noteworthy that both \(W_0\) and \(\Delta W = BA\) are applied to the same input, with their outputs being aggregated coordinate-wise. 
For an output \(h = W_0x\), the modified forward pass is:
\begin{equation}
h = W_0x + \Delta Wx = W_0x + BAx
\end{equation}
Typically, A is initialized with a random Gaussian distribution, and B with zero, ensuring that \(\Delta W = BA\) starts from zero at the inception of training. The scaling of \(\Delta Wx\) by \(\alpha/r\). The rank of the low rank matrices is denoted by \(r\),  whereas $\alpha$ is the scaling factor that controls the magnitude of updates to the matrices. 

% , where \(\alpha\) is a predefined constant in \(r\), further delineates this reparametrization strategy. 

\subsection{Evaluation Metrics}
\begin{table*}[t!]
\centering
\caption{Table comparing different variations of Video LLaVA and LLaVA NeXT Video models (Base, LoRA, QLoRA, and Pruning) using ROUGE-1, ROUGE-2, ROUGE-L, BLEU, and BERTScore metrics.}
\resizebox{\textwidth}{!}{%
\begin{tabular}{l c c c c c c c c c c c}
\toprule
                   & \multicolumn{5}{c}{\textbf{Video LLaVA 7B}} && \multicolumn{5}{c}{\textbf{LLaVA NeXT Video 7B}} 
                    \\ \cmidrule(lr){2-6} \cmidrule(lr){8-12} 
                   & ROUGE-1
                   & ROUGE-2
                   & ROUGE-L
                   & BLEU & BERT & &ROUGE-1 & ROUGE-2 & ROUGE-L & BLEU & BERT \\ 
                    \midrule

\textbf{Base}     & 0.211 & 0.041 & 0.122 & 0.039 & 0.456 && 0.197 & 0.037 & 0.113 & 0.042 & 0.404 \\

\textbf{10K LORA}     & 0.563 & 0.214 & 0.313 & 0.243 & 0.849 && 0.554 & 0.198 & 0.300 & 0.232 & 0.856 \\ 

\textbf{100K LORA}     & \textbf{0.576} & \textbf{0.226} & \textbf{0.325} & \textbf{0.250} & \textbf{0.863} && \textbf{0.562} & 0.199 & 0.300 & \textbf{0.239} & \textbf{0.864} \\

\textbf{10K-QLORA}     & 0.565 & 0.212 & 0.310 & 0.243 & 0.845 && 0.543 & 0.193 & 0.283 & 0.213 & 0.836 \\ 

\textbf{100K-QLORA}     & 0.571 & 0.220 & 0.316 & \textbf{0.250} & 0.854 && 0.561 & \textbf{0.202} & \textbf{0.302} & 0.229 & 0.858 \\

\textbf{10K Pruning\_{5\%}}     & 0.031 & 0.007 & 0.024 & 0.010 & 0.265 && 0.532 & 0.178 & 0.278 & 0.209 & 0.829 \\ 

\textbf{100K Pruning\_{5\%}}     & 0.125 & 0.034 & 0.110 & 0.043 & 0.359 && 0.541 & 0.183 & 0.284 & 0.210 & 0.840 \\ 

\textbf{Final Model}     & - & - & - & - & - && \underline{0.556} & \underline{0.202} & \underline{0.290} & \underline{0.227} & \underline{0.850} \\
\bottomrule
\end{tabular}%
}
\label{main}
\vspace{0.5em}
\end{table*}

Many evaluation metrics are taken into consideration to evaluate the performance of the fine-tuned models and evaluate the model's generated text against the ground truth text. In similar works about fine-tuned models in domain-specific tasks, metrics such as ROUGE, BLEU, and METEOR are used. All three metrics compare the overlap of n-grams or phrases between the generated output and the reference text. 

\textbf{ROUGE Score} \cite{lin2004rouge} measures the N-gram overlap between a candidate’s output and a set of reference outputs, ROUGE-1, ROUGE-2, and ROUGE-L were used. Where 1 and 2 are the n-grams, and ROUGE-L is based on the \textbf{L}ongest common subsequence. 

\textbf{BLEU} \cite{papineni2002bleu} is commonly used for generation tasks to measure n-gram similarities between machine-generated outputs and reference translations. Although our work is not focused on translation, we use this metric to assess our generated outputs. It is calculated as follows:

\begin{equation}
\text{BLEU} = \text{BP} \cdot \exp \left( \sum_{n=1}^N w_n \log p_n \right)
\end{equation}

where $p_n$ is the precision for n-grams of length $n$, $w_n$ are weights (50\% for each n-gram in this work), and $N$ is the maximum n-gram length considered, $N=2$ in this work.

\vspace{0.5cm}
Although these two metrics give a good idea of the model's performance by comparing words and sentences and matching them against the reference text, they both have limited contextual understanding or capture the semantic coherence of the generated or reference texts. Therefore, the BERT metric is also utilized to provide performance indicators between texts by generating contextual embedding to capture the semantic similarity between words. 

\textbf{BERT Score} \cite{zhang2019bertscore} employs BERT \cite{devlin2018bert} to assess similarity between two sentences by aligning each token in the reference sentence with the closest token in the candidate sentence. The similarity is determined through the cosine similarity of token embeddings. Precision is calculated by comparing the candidate tokens with those in the reference, while recall involves matching reference tokens to those in the candidate. The F1 score is subsequently derived from both precision and recall. The formulas for recall, precision, and F1 are:

\begin{equation}
R_{\text{BERT}} = \frac{1}{|x|} \sum_{x_i \in x} \max_{x_j \in \hat{x}} x_i^T \hat{x}_j
\end{equation}

\begin{equation}
P_{\text{BERT}} = \frac{1}{|\hat{x}|} \sum_{\hat{x}_j \in \hat{x}} \max_{x_i \in x} x_i^T \hat{x}_j
\end{equation}

\begin{equation}
F_{\text{BERT}} = \frac{2 \cdot P_{\text{BERT}} \cdot R_{\text{BERT}}}{P_{\text{BERT}} + R_{\text{BERT}}}
\end{equation}

where $x$ and $\hat{x}$ denote the reference and candidate sentences, respectively, and $x_i^T \hat{x}_j$ represents the cosine similarity between token embeddings $x_i$ and $\hat{x}_j$.

\begin{table*}[t!]
\centering
\caption{
Table comparing various configurations of LLaVA and LLaVA NeXT models (including QLoRA with 4-bit and 8-bit precision and Pruning at 10$\%$) across evaluation metrics: ROUGE-1, ROUGE-2, ROUGE-L, BLEU, and BERTScore. The variations are analyzed using different ranks (r) and alpha values ($\alpha$).}
\resizebox{\textwidth}{!}{%
\begin{tabular}{l c c c c c c c c c c c c c}
\toprule
                   & &&\multicolumn{5}{c}{\textbf{Video LLaVA 7B}} && \multicolumn{5}{c}{\textbf{LLaVA NeXT Video 7B}} 
                    \\ \cmidrule(lr){4-8} \cmidrule(lr){10-14} 
                   &r &$\alpha$& ROUGE-1  & ROUGE-2 & ROUGE-L & BLEU & BERT & &ROUGE-1 & ROUGE-2 & ROUGE-L & BLEU & BERT \\ 
                    \midrule

\textbf{10K QLORA-4bit}     &640 &1280&0.561	&0.211	&0.313	&0.236	&0.855&		&0.531	&0.177	&0.281	&0.196	&0.834 \\

\textbf{100K QLORA-4bit}    &640 &1280& 0.570&	0.219&	0.318&	0.245&	0.860&&0.545&0.187  & 0.287& 0.205&0.844\\ 

\textbf{10K QLORA-4bit}    &64 &256 & 0.556	&0.203	&0.307	&0.233	&0.854 &&0.539	&0.185	&0.288	&0.213	&0.834\\

\textbf{100K QLORA-4bit}   &64& 256  &0.567&	0.212&	0.311	&0.240&	0.863&&0.555&	\textbf{0.196}&	\textbf{0.296}&	\textbf{0.225}&	0.848 \\ 

\textbf{10K QLORA-8bit}    &64 &128&0.566&	0.221&	0.319&	0.248&	0.844	&&0.542	&0.183	&0.288	&0.211	&0.842\\ 

\textbf{100K QLORA-8bit}   &64 &128 & \textbf{0.578}	&\textbf{0.224}	&\textbf{0.320}	&\textbf{0.252}	&\textbf{0.863}&		&\textbf{0.557}	&0.195	&0.294	&0.224	&\textbf{0.852} \\ 

\textbf{10K Pruning\_{10\%}}    &64& 128 & 0.025 & 0.005 & 0.018 & 0.008 & 0.250 && 0.479&	0.130&	0.224&	0.171&	0.747 \\ 

\textbf{100K Pruning\_{10\%}}  &64 &128   & 0.063 & 0.018 & 0.052 & 0.021 & 0.289 && 0.529&0.175&0.274&0.204&0.823\\ 

\bottomrule
\end{tabular}%
}
\label{abalation}
\end{table*}

\begin{figure*}[t!]
    \centering
    \begin{subfigure}{0.48\textwidth}
        \centering
        \includegraphics[width=\linewidth]{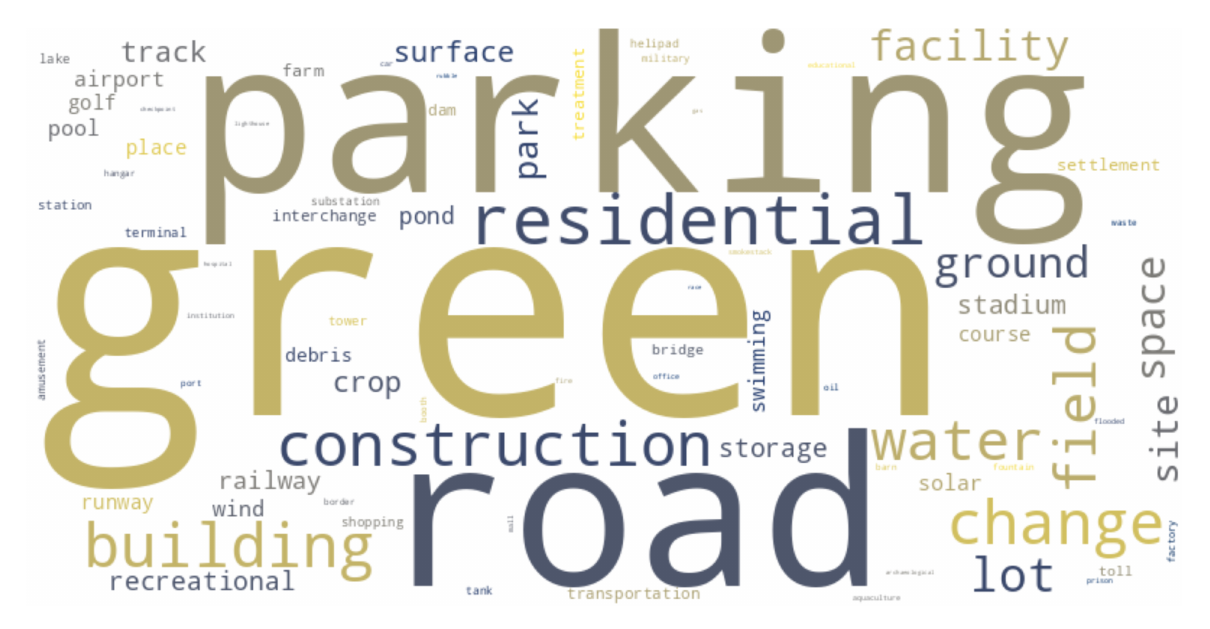}
        \caption{Ground truth captions.}
        \label{fig:image1}
    \end{subfigure}
    \hfill
    \begin{subfigure}{0.48\textwidth}
        \centering
        \includegraphics[width=\linewidth]{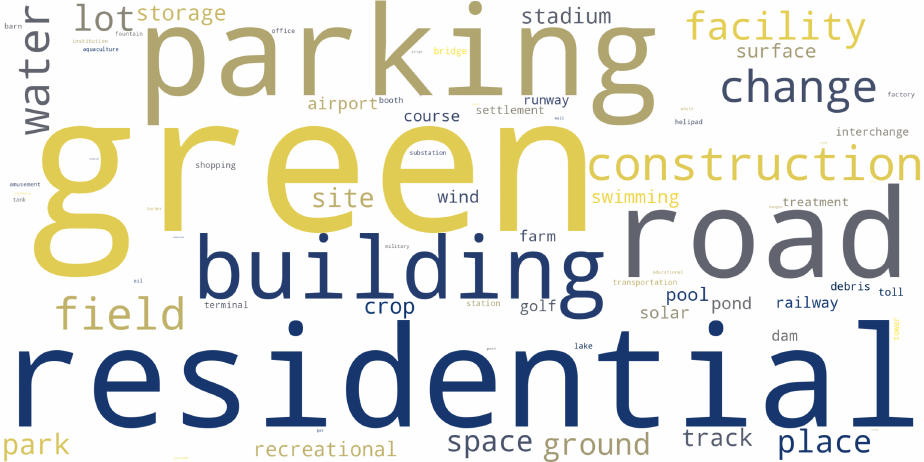}
        \caption{Final model's generated captions.}
        \label{fig:image2}
    \end{subfigure}
    \vspace{1.5em}
    \caption{Word clouds comparing the ground truth (left) and the final model's generated annotations (right).}
    \label{fig:combined}
    \vspace{0.5em}
\end{figure*}

\subsection{Model Optimization}
\textbf{Pruning}: We implement magnitude-based fine-grained pruning, an unstructured pruning method that selectively removes individual weights based on their magnitude. Precisely, the weights with the smallest absolute values are pruned for each layer according to a predefined sparsity level. The pruning process uses a binary mask that retains important weights (those with larger magnitudes) and sets the others to zero. A global sparsity target is applied to ensure consistent pruning across the model, although certain layers, such as embeddings and critical vision model components, are excluded to preserve model performance. This fine-grained approach allows for more granular control over which weights are pruned, resulting in reduced model size and computational overhead with minimal impact on accuracy. During inference, the pruning masks are re-applied to maintain the enforced sparsity, optimizing the model for efficiency without sacrificing performance.

In magnitude-based fine-grained pruning, each weight \( W_{i,j} \) in the weight matrix \( W \) is pruned based on its absolute value \( |W_{i,j}| \). The pruning threshold \( \tau \) is determined such that the smallest \( s \times 100\% \) of weights, where \( s \) is the sparsity level, are pruned. A binary mask \( M \) is created, where

\begin{equation}
M_{i,j} = 
\begin{cases} 
1 & \text{if} \ |W_{i,j}| > \tau \\
0 & \text{if} \ |W_{i,j}| \leq \tau
\end{cases}
\end{equation}

The pruned weights are then obtained by element-wise multiplication of the original weight matrix \( W \) with the mask \( M \), yielding

\begin{equation}
W^{\text{pruned}} = W \odot M
\end{equation}

This process reduces the model's parameter count while retaining the most significant weights.

\textbf{QLORA}: Although LoRA reduces the overall number of parameters to be modified from the original model, it is still challenging to fine-tune the total number on a single device. Therefore, quantization for LLMs was introduced in \cite{qlora} to optimize computation for reduced memory usage while maintaining model accuracy, referred to as QLORA. The quantization (q) is calculated by:

\begin{equation}
q = \text{round}\left(\frac{(2^b - 1)}{\text{absmax}(X)} \cdot X\right) = \text{round}(c \cdot X)
\end{equation}

where \( c \) is the quantization constant or quantization scale. 

\vspace{0.5cm}
Ultimately, all models were fine-tuned with a single 48GB A6000 GPU, for one epoch, taking on average between 2 hours and 24 hours with batch size 3 for Video-LLaVA 7B for the 10k and 100k datasets, respectively. As for LLaVA-NeXT Video 7B, the batch size was 2, tuning for 3 to 27 hours for the 10k and 100k datasets, respectively. The final model's full hyperparameters and training configurations can be found in Table \ref{tab:hyperparams}.

\begin{table*}[h!]
\centering
\caption{Full hyper-parameters used for fine-tuning the final model.}
\begin{tabular}{|lc|lc|lc|}
\hline
\textbf{Parameter}  & \textbf{Value} & \textbf{Parameter}        & \textbf{Value}         & \textbf{Parameter}          & \textbf{Value} \\
MAX LENGTH          & 400            & MODEL                     & LLaVA-NeXT-Video-7B-hf & USE\_QLORA                  & True (4-Bit)   \\
batch\_size         & 2              & lora\_r                   & 64                     & lora\_alpha                 & 128            \\
max\_epochs         & 1              & val\_check\_interval      & 0.2                    & check\_val\_every\_n\_epoch & 1              \\
gradient\_clip\_val & 1.0            & accumulate\_grad\_batches & 1                      & learning\_rate              & 1e-4           \\
num\_nodes          & 1              & warmup\_steps             & 50                     &                             &                \\ \hline
\end{tabular}
\label{tab:hyperparams}
\end{table*}

\section{Results \& Discussion}

Table \ref{main} presents results across various models and scoring metrics. All experiments were run on a single 40GB GPU, optimizing LoRA configurations and pruning ratios for a balance of accuracy and memory efficiency. Although larger VLMs may perform better, they require substantially more computational resources.

Since prior work lacked image descriptions or did not employ the fMoW dataset for captioning, we evaluate Video-LLaVA and LLaVA-NeXT-Video as zero-shot baselines. Without fine-tuning, these models performed poorly across all metrics and failed to capture meaningful semantic differences, even under BERT-based evaluation.

We then applied few-shot fine-tuning with 10K (10\%) and 100K samples using LoRA ($r=64$, $\alpha=128$), tuning 178M parameters. Performance improved notably, with the 100K setup achieving a BERT score of 0.864.

To enhance efficiency, QLoRA with 4-bit quantization reduced memory usage by ~75\% without sacrificing accuracy, achieving a BLEU score of 0.250, comparable to the LoRA-based model.

\begin{figure*}[t!]
    \centering
    \includegraphics[width=0.79\textwidth]{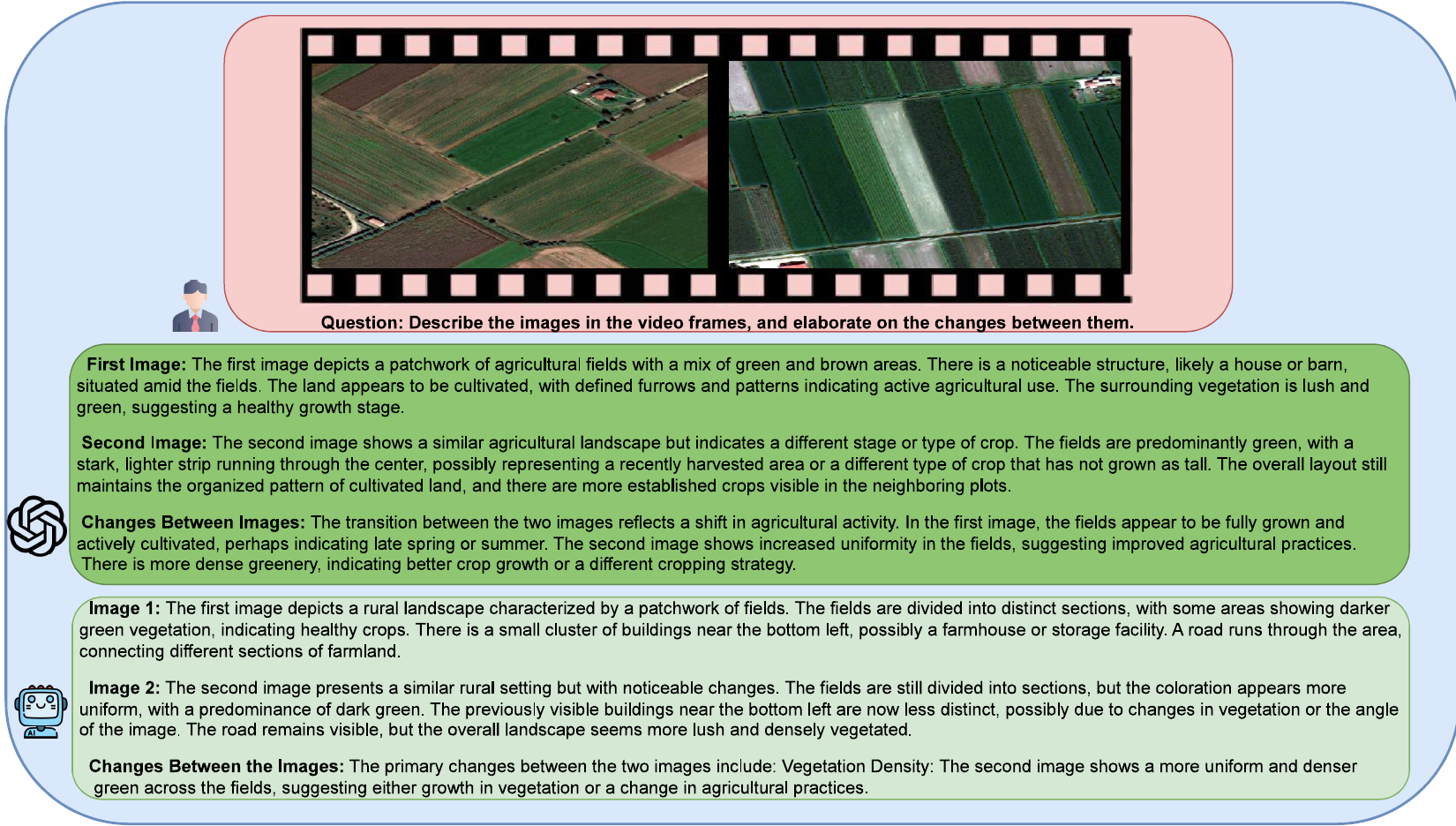}
    \vspace{0.5em}
    \caption{Qualitative output comparing the output from ChatGPT vs our model's for two images that look similar but for different locations, showcasing the ability of distinguishing differences}
    \vspace{0.5em}
    \label{fig:qualitative_comparison}
\end{figure*}

\begin{figure*}[t!]
    \centering
    \includegraphics[width=0.69\textwidth]{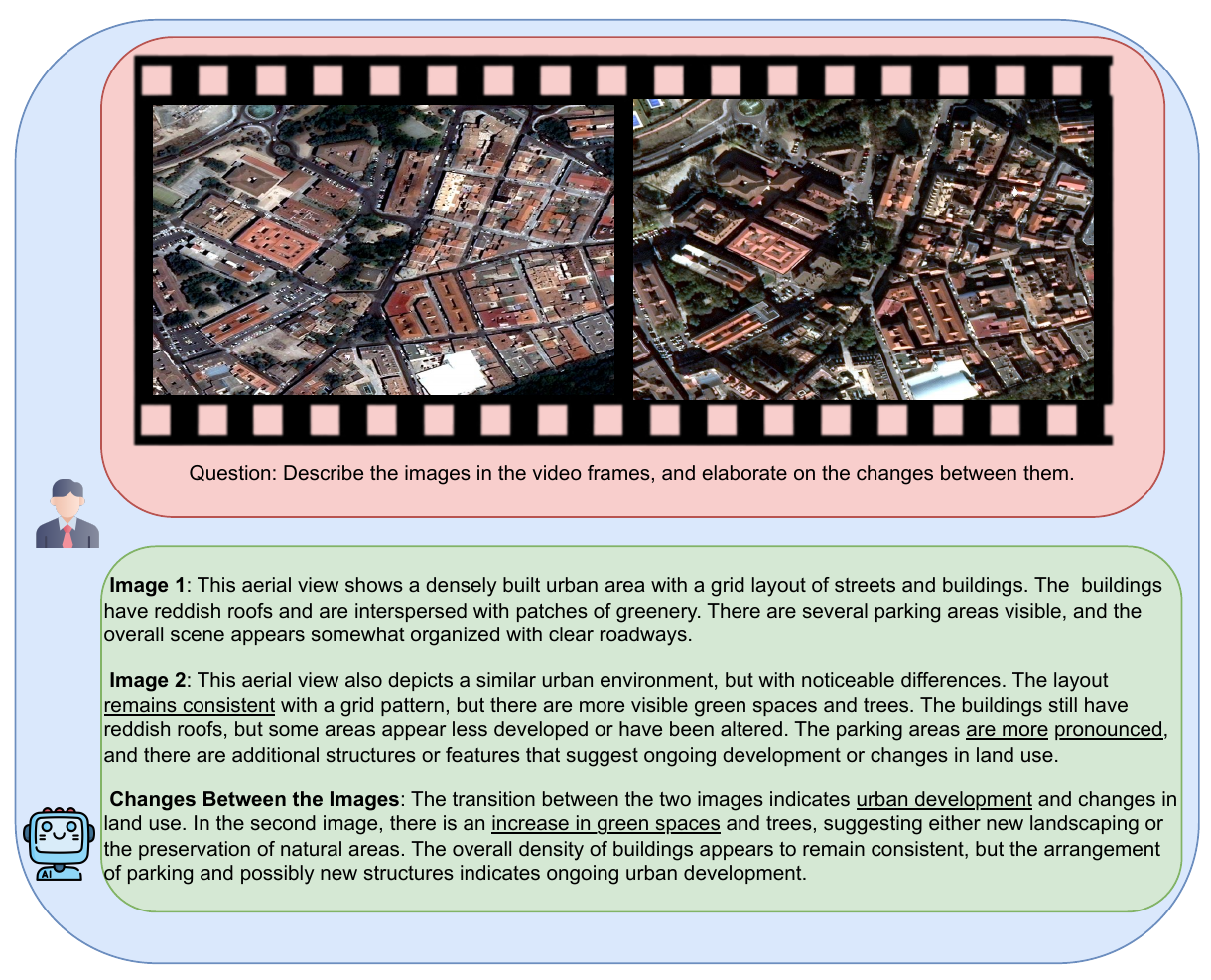}
    \vspace{0.5em}
    \caption{Qualitative output showcasing a sample video of two frames inputted with a question followed with the model output describing the two images and summarizing differences and changes.}
    \vspace{0.5em}
    \label{fig:qualitative}
\end{figure*}

We pruned 5\% of the model parameters to reduce size while preserving accuracy. However, pruning widened the performance gap between the 10K and 100K datasets, indicating a greater data requirement to offset pruning-induced degradation. Notably, the LLaVA-Next video model outperformed Video-LLaVA, thanks to its sparse structure, achieving a BERT score of 0.823—only 0.03 lower than the best result. In contrast, Video-LLaVA faced significant challenges with pruning due to its dense architecture, making it unsuitable for pruning.

Table \ref{abalation} summarizes our ablation study to validate the chosen hyperparameters. We explored several $\alpha$ and $r$ ratios to determine their impact on performance. Increasing the fine-tuned parameters to 1.7B by setting $r=640$ and $\alpha=1280$ did not yield significant performance gains, highlighting diminishing returns at higher parameter counts. Modifying the $\alpha$ and $r$ ratio to 4:1 also resulted in negligible improvements or degraded performance. Therefore, we adopted the $r=64$ and $\alpha=128$ configuration for subsequent experiments.

Although 8-bit quantization slightly improved performance, it increased memory usage by 150\% and fine-tuning time by 174\% compared to the 4-bit model. Similarly, increasing the pruning to 10\% reduced the size but significantly reduced the accuracy. Thus, we adopted 4-bit QLoRA with 5\% pruning as a balanced configuration. While greater pruning might suit applications prioritizing efficiency over accuracy, we chose the intermediate solution of 5\% pruning to balance performance and efficiency.

Using 4-bit QLoRA, 5\%  pruning, and fine-tuning on 100K samples, our final model demonstrated competitive performance, achieving a BERT score of 0.850. The results show that although this setup does not yield the highest possible scores, it strikes an optimal balance between accuracy and efficiency. Specifically, the model is 5\%  less resource-intensive while still performing well.

For qualitative analysis, Figure \ref{fig:combined} presents word cloud representations comparing the ground truth and the model-generated captions. The visual overlap highlights the alignment between key descriptive terms in both the ground truth and the model outputs, showcasing the model’s ability to capture salient features. Figure  \ref{fig:qualitative} provides an additional qualitative example, where the model describes two input images with high attention to key objects. The model accurately identifies changes between the images, using smooth and coherent language to summarize differences.

Additionally, we evaluated the model's ability to detect changes between image pairs that, while visually similar, originate from different geographic locations—posing a potential challenge to its change detection capabilities. Despite this, the model demonstrated robust performance, effectively describing each image and identifying key differences. Its responses were semantically comparable to those generated by GPT, as illustrated in Figure~\ref{fig:qualitative_comparison}.

Our results align closely with SOTA models such as GPT-4, and the generated annotations demonstrate strong consistency with human descriptions.

\section{Conclusion}
In this paper, we introduced a novel dataset and applied fine-tuning techniques with it to enhance VLMs for detecting temporal changes in geographical landscapes. By employing methods such as LoRA and QLoRA, we enhanced models like Video-LLaVA and LLaVA-NeXT-Video. Our fine-tuned models surpassed the performance of the base models, with the final model achieving a BERT score of 0.864 and a ROUGE-1 score of 0.576. Furthermore, the use of quantization and pruning improved computational efficiency without degrading accuracy, making the models more suitable for real-time applications. This work addresses key limitations in remote sensing, providing a scalable and efficient solution for tracking temporal changes in environmental and urban landscapes.

\section{Limitation \& Future Work}

\textbf{Reliance on GPT-4o Mini for Annotations:} 
We relied on a single model for annotations, providing reliable ground truth. However, leveraging other models could offer a broader comparison and capture more nuances in temporal changes. Future work could integrate multiple models to improve annotation diversity and robustness.

\textbf{Limited Dataset Due to Labeling Costs:}
Labeling with GPT-4o mini limited us to 100,000 image pairs for training and 6,000 for testing. Larger datasets would provide diverse examples, improving fine-tuning and generalization. Future efforts should prioritize creating larger datasets to enhance model training and performance.

\textbf{Single Dataset:}
Our dataset, while diverse in classes, locations, and image characteristics, is still singular. Using datasets with varied collection schemas could improve generalization. Future work should include multiple data sources with different temporal resolutions to expand generalization.

\textbf{Manual Evaluation and Crowdsourcing:}
Manual evaluation was restricted to test data due to the training data volume. Crowdsourcing could enhance verification and quality but increases costs. Future work could explore crowdsourcing to validate and improve dataset quality at scale.

\textbf{Hardware Limitations:}
We used a single GPU, limiting processing speed and batch sizes. Access to more advanced GPUs would enable faster training, larger batches, and more complex models. Future research should leverage better hardware for training efficiency.

\textbf{Supervised Learning Only:}
We used supervised learning, which provided reliable results but limited generalization to unseen data. Future work could explore unsupervised or semi-supervised approaches for more robust results, especially in low-resource settings.

\textbf{Model Distillation:}
Teacher-student distillation techniques were not explored. Distillation could reduce model complexity and computational needs while maintaining performance levels. Future research could focus on distillation to make models more efficient for real-world applications.

%%%%%%%%%%%%%%%%%%%%%%%%%%%%%%%%%%%%%%%%%%%%%%%%%%%%%%%%%%%%%%%%%%%%%%%%

%%% Use this command to include your bibliography file.

\bibliography{mybibfile}

\end{document}